\documentclass[runningheads]{llncs}
\usepackage{graphicx}

\usepackage{tikz}
\usepackage{comment}
\usepackage{amsmath,amssymb} 
\usepackage{color}
\usepackage{bbding}

\usepackage[accsupp]{axessibility}  

\usepackage{cite}
\usepackage{mathrsfs}
\usepackage{xcolor}
\usepackage{comment}
\usepackage{pifont}

\usepackage{color}
\usepackage{hyperref}
\definecolor{myGreen}{RGB}{55,149,73}

\usepackage{setspace}
\usepackage{amsmath}
\usepackage[ruled,linesnumbered]{algorithm2e}
\usepackage{multirow}
\usepackage{caption}

\begin{document}
\pagestyle{headings}
\mainmatter
\def\ECCVSubNumber{0000}  

\title{EAN-MapNet: Efficient Vectorized HD Map Construction with Anchor Neighborhoods} 

\titlerunning{EAN-MapNet}
%
\author{
	Huiyuan Xiong \inst{1,2}
	Jun Shen \inst{1,2}
	 Taohong Zhu \inst{1,2}
	Yuelong Pan\inst{3,4} \thanks{Corresponding author.}
}
%
%
\institute{
	School of Intelligent Systems Engineering, Sun Yat-sen University \\
	\email{\{shenj57, zhuth3\}@mail2.sysu.edu.cn} \\
	\email{xionghy@mail.sysu.edu.cn} \and
	Guangdong Provincial Key Laboratory of Intelligent Transport System \and
	China Nuclear Power Design Co.,Ltd. \\
	\email{panyuelong@cgnpc.com.cn} \and
	State Key Laboratory of Nuclear Power Safety Technology and Equipment, China Nuclear Power Engineering Co., Ltd., Shenzhen, Guangdong, 518172, China
}
\maketitle

\begin{abstract}
	High-definition (HD) map is crucial for autonomous driving systems.
	Most existing works design map elements detection heads based on the DETR decoder.
	However, the initial queries lack explicit incorporation of physical positional information, and vanilla self-attention entails high computational complexity.
	Therefore, we propose {\bf{EAN-MapNet}} for {\bf{E}}fficiently constructing HD map using {\bf{A}}nchor {\bf{N}}eighborhoods. Firstly, we design query units based on the anchor neighborhoods, allowing non-neighborhood central anchors to effectively assist in fitting the neighborhood central anchors to the target points representing map elements. 
	Then, we propose grouped local self-attention ({\bf{GL-SA}}) by leveraging the relative instance relationship among the queries. This facilitates direct feature interaction among queries of the same instances, while innovatively employing local queries as intermediaries for interaction among queries from different instances. Consequently, GL-SA significantly reduces the computational complexity of self-attention while ensuring ample feature interaction among queries.
	On the nuScenes dataset, EAN-MapNet achieves a state-of-the-art performance with 63.0 mAP after training for 24 epochs, surpassing MapTR by 12.7 mAP. Furthermore, it considerably reduces memory consumption by 8198M compared to MapTRv2.
\end{abstract}

\section{Introduction}
\label{sec:intro}
%
In autonomous driving systems, HD map leverage rich semantic information, accurately reflecting the road environment, and providing robust data support for planning and decision-making \cite{bansal2018chauffeurnet,cui2019multimodal,chai2019multipath}. However, the traditional creation process of HD map begins with capturing point clouds and constructing map by SLAM \cite{zhang2014loam,mur2017orb}. Subsequently, it heavily relies on a large amount of human resources for annotation and maintenance. Consequently, traditional method faces challenges such as high construction costs, poor scalability, and low data freshness.

In recent years, the researchs on end-to-end HD map construction have gained significant attention. Early works focus on semantic segmentation of the bird's eye view (BEV) to identify map elements pixel by pixel\cite{li2022hdmapnet,dong2022superfusion,peng2023bevsegformer}. However, it is evident that these works require intricate post-processing steps to generate the vectorized representation of map elements.
Therefore, the approaches of representing map elements as polylines composed of point sequences have been introduced\cite{liao2022maptr,liao2023maptrv2,yuan2024streammapnet,xu2023insightmapper,yu2023scalablemap}.
Typically, such methods design detection heads based on the DETR decoder\cite{carion2020end}.
The initial query aggregates the feature, updating the coordinates of the reference points, thus continuously fitting them to the target points

\begin{figure}[tb]
	\centering
	\includegraphics[width=1\textwidth, trim=15cm 9cm 10cm 7cm, clip]{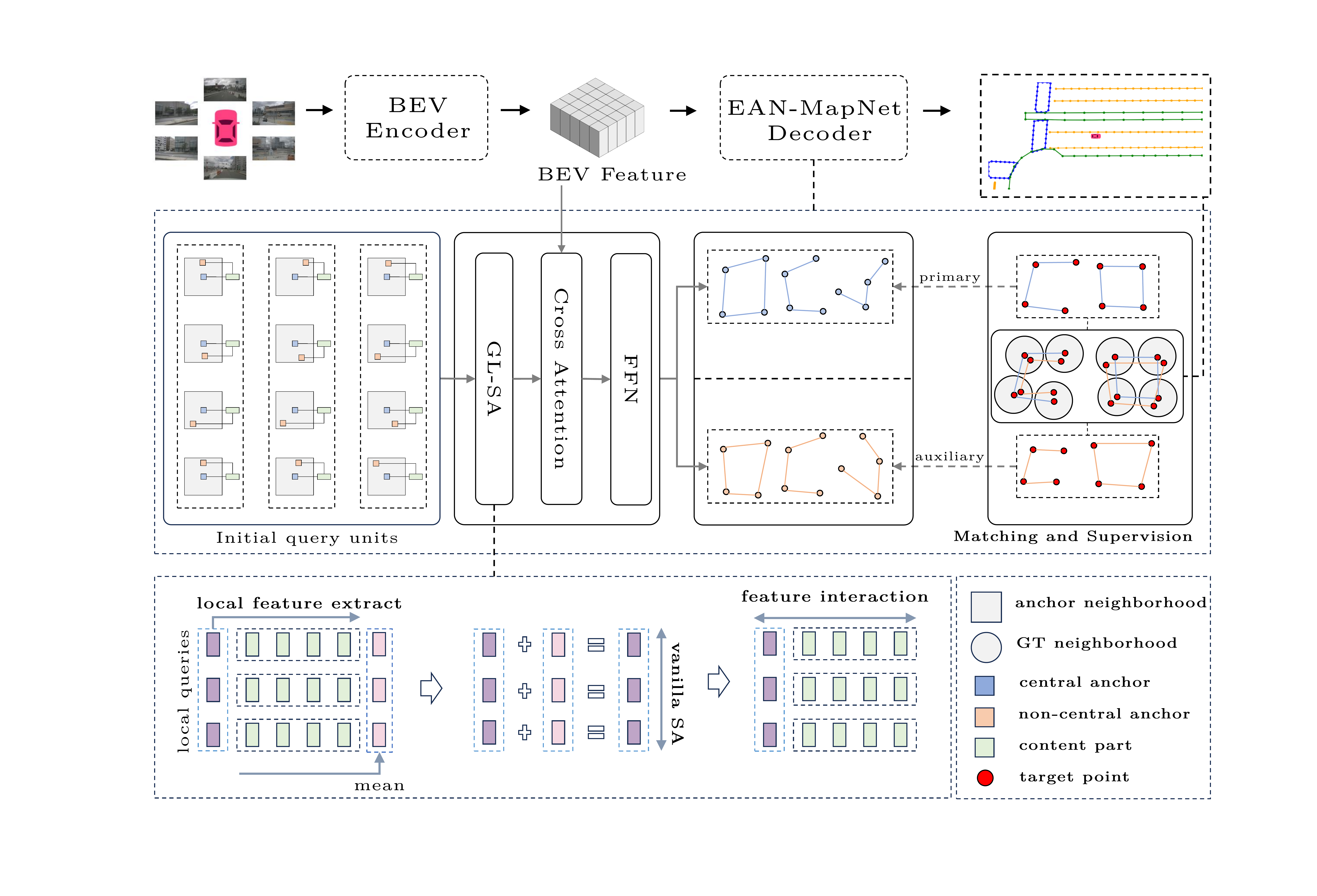}
	\caption{
		The overall architecture of EAN-MapNet: First, images captured by the surround-view cameras undergo transformation into unified BEV feature by the BEV Encoder.  
		Then, in the decoder, the auxiliary part assists the primary part in updating the neighborhood central anchors to the target points by fitting the non-neighborhood central anchors into the ground truth(GT) neighborhoods.
		GL-SA initially extracts local features of anchor queries within each group using local queries. Subsequently, vanilla self-attention is applied to the local queries, efficiently enabling feature interaction among groups. Then, each local query is assigned to its corresponding group of original query units, followed by ample feature interaction within each group.
	}
	\label{fig1}
\end{figure}

However, these approaches have the following two drawbacks: One drawback is the design of queries neglects the incorporation of physical positional information. So, the coordinates of the initial reference points are predicted by a multilayer perceptron (MLP) from $d$-dimensional positional queries. However, since the positional queries are unrelated to BEV, they restrict the networks' comprehension of positional prior information\cite{li2023lanesegnet} , such as the explicit relative position encoding. 
Another drawback is the high computational complexity of vanilla self-attention, making it challenging to meet the demand for setting a large number of queries to enhance model prediction diversity. Furthermore, vanilla self-attention overlooks the strong feature correlation among queries within the same instances.

Therefore, we propose EAN-MapNet, whose overall structure is illustrated in \textcolor{red}{Fig}\eqref{fig1}. We randomly initialize a certain number of anchors. The task of detecting map elements is formulated as the process of fitting initial anchors to target points.

\begin{figure}[tb]
	\centering
	\includegraphics[width=1\textwidth, trim=2cm 4cm 4cm 4cm, clip]{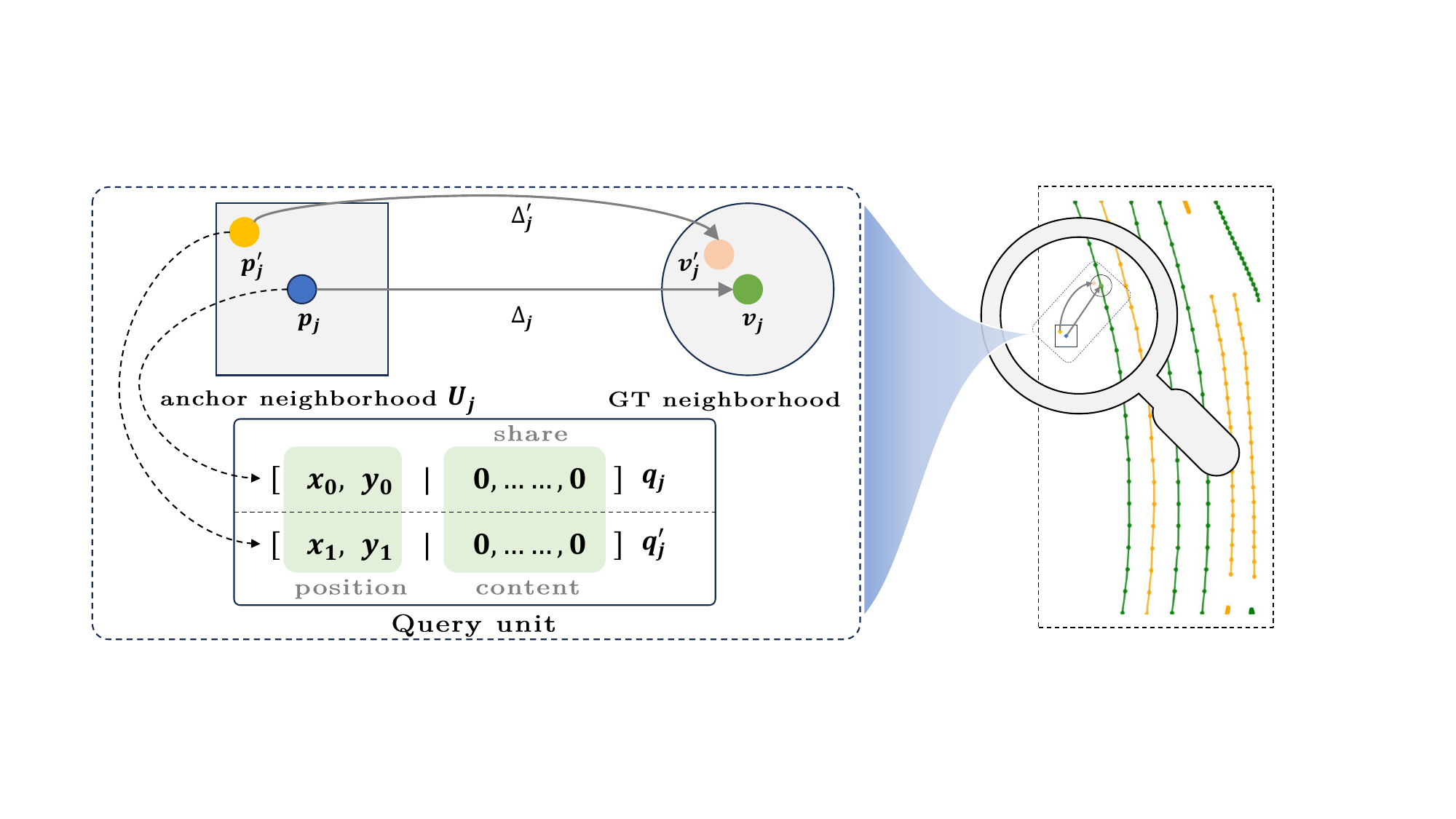}
	\caption{
		A single query unit corresponds to two queries, each composed of the $2$-dimensional coordinates of anchors situated in the initial neighborhood, along with shared $d$-dimensional learnable parameters. The neighborhood central anchor is fitted to the target point, while the non-neighborhood central anchor is fitted to a random point within the GT neighborhood.
	}
	\label{fig2}
\end{figure}

In order to more effectively leverage the physical positional information of anchors, we define a neighborhood around each anchor, with each neighborhood corresponding to a query unit.
As illustrated in \textcolor{red}{Fig}\eqref{fig2}, a query unit consists of a neighborhood central query and a non-neighborhood central query, each comprising positional and content parts. 
Non-neighborhood central anchors and neighborhood central anchors exhibit strong correlations in terms of positional prior information. So, during training, they are allowed to share content parts, enabling non-neighborhood central anchors to effectively assist in fitting the neighborhood central anchors to the target points.
It is important to note that neighborhood central anchors are fitted to target points, while non-neighborhood central anchors are fitted to random points within the GT neighborhoods, thereby mitigating fitting ambiguities. Additionally, the non-neighborhood central anchor queries are exclusively involved in the network training process and do not consume computational resources required for network inference.

Queries within the same groups correspond to vertices of the same map elements, exhibiting strong feature correlation among them. Meanwhile, the relative positional relationship among vertices in different groups is also crucial information. 
Therefore, we propose GL-SA, where each query interacts only with other queries within the same groups, while interactions with queries from different groups are mediated by local queries. 
Compared to vanilla self-attention, where each query interacts with all queries, GL-SA significantly reduces computational complexity while enabling ample feature interaction among queries.

The contributions of this paper are as follows:
\begin{itemize}
	\item{We formulate query units that incorporate the physical positional information of the anchor neighborhoods, leveraging non-neighborhood central anchors to effectively assist in fitting the neighborhood central anchors to the target points.}
	\item{We propose GL-SA, which enhances feature interaction among queries within the same groups and efficiently facilitates feature interaction among queries from different groups.}
	\item{We introduce EAN-MapNet, which achieves a 63.0 mAP after training for 24 epochs, outperforming existing state-of-the-art approaches, while maintaining low memory consumption.}
	
\end{itemize}

\section{Related Works}

\subsection{DETR}
The structure of the map elements detection heads in EAN-MapNet is inspired by the DETR decoder. However, DETR suffers from slow convergence, prompting numerous approaches to address this issue. 
Deformable DETR\cite{zhu2020deformable} introduces deformable attention. each query is updated based on information extracted from sampled points on the feature maps output by the encoder. 
DN-DETR\cite{li2022dn} and DION\cite{zhang2022dino} incorporate a denoising part into network training, which takes GT labels with randomly added noise as input and is directly supervised by GT labels, bypassing Hungarian matching to mitigate ambiguity from unstable matching. The inspiration for the neighborhood in EAN-MapNet is derived from the denoising part. 
Anchor DETR\cite{wang2021anchor}, Conditional DETR\cite{meng2021conditional}, and DAB-DETR\cite{liu2022dab} initialize queries based on anchor coordinates, thereby providing them with explicit physical meaning. However, anchor queries only incorporate the position information of the anchors themselves, neglecting the positional information of the surrounding anchor regions. Therefore, we propose designing query units based on anchor neighborhoods to address this limitation.

\subsection{Self-Attention}
In vanilla self-attention\cite{vaswani2017attention}, each query interacts with all other queries. To improve computational efficiency, methods like multi-query attention\cite{shazeer2019fast} and grouped-query attention \cite{ainslie2023gqa} adopt a strategy of sharing keys and values among queries. Sparse Transformer\cite{child2019generating} defines factorized attention for rows and columns, enabling sparse decomposition of the attention matrix. Additionally, due to the sparsity of self-attention, Informer\cite{zhou2021informer} introduces probsparse self-attention, which filters queries based on their sparsity measure to reduce computational complexity. 
However, these attention mechanisms struggle to effectively leverage the grouped nature of queries to achieve efficient and ample feature interaction. Therefore, we introduce GL-SA. GL-SA strengthens feature interaction among queries within the same groups and facilitates interactions among queries from different groups using local queries.

\subsection{Representation of Map Elements}
To achieve end-to-end HD map construction, the first step is to determine the representation of map elements. The pixel-based representation methods\cite{li2022hdmapnet,dong2022superfusion,gao2023complementing,xie2023mv} rasterize map elements into pixels and perform pixel-level classification on the BEV to identify map elements. These methods require complex post-processing to generate vectorized representation of map elements. Therefore, MapTR\cite{liao2022maptr}, StreamMapNet\cite{yuan2024streammapnet} and InsightMapper\cite{xu2023insightmapper} describe map elements as equivalent point sequences. The initial queries fit the reference points to target points by aggregating features. EAN-MapNet follows a similar approach. However, ambiguities arise if anchors within the same anchor neighborhoods are fitted to the same target points. Thus, we propose reconstructing map elements based on GT neighborhoods to obtain the fitting targets for non-neighborhood central anchors. 
In other words, each map element is represented as a combination of the original polyline and the reconstructed polyline.

\section{Methods}

\subsection{Grouped Anchor Query Units}
Firstly, we represent map elements as a collection of polylines, denoted as $ V = \left \{ V_{i} \right \}_{i=0}^{M} $, where $ M $ denotes the number of instances of map elements. Each polyline $ V_{i} $ is composed of an ordered set of vertices, denoted as  $ V_{i} = \left \{ v_{j} \right \}_{j=0}^{N} $, where $ N $ denotes  the number of vertices in each map element.

Then, we randomly initialize a group of anchors on the BEV plane, denoted as $ \left \{ p_{j} \right \}_{j=0}^{N} $. Subsequently, a square neighborhood with side length $a$ is created around each anchor, and non-central anchors are denoted as  $ \left \{ p_{j}' \right \}_{j=0}^{N} $.
As illustrated in \textcolor{red}{Fig}\eqref{fig2}, considering an anchor neighborhood $U_{j}$, we respectively construct a neighborhood central anchor query $q_{j}$ and a non-neighborhood central anchor query $q_{j}'$ for initial anchors $p_{j}$ and $p_{j}'$. $q_{j}$ and $q_{j}'$ together form a query unit, each of which can be decomposed into positional and content parts, as shown in \textcolor{red}{Equation}\eqref{pythagorean1}.
Here, the content part $ c_{j} $ is shared and represented as an $d$-dimensional learnable parameter responsible for generating the offsets $ \Delta_{j} $ and $ \Delta_{j}' $  from the anchors to the target points, along with the categorical information of the anchors.

\begin{equation}
	\left\{\begin{matrix}
		& q_{j} = Cat(p_{j},c_{j})\\ 
		& q_{j}' = Cat(p_{j}',c_{j})\\
	\end{matrix}\right. 
	\label{pythagorean1},
\end{equation}
where $Cat(.)$ represents concatenation on the embedding dimension.

A group of query units can represent only a single map element. Inspired by the hierarchical query embedding scheme of MapTR\cite{liao2022maptr}, we introduce grouping embeddings for the positional and content parts, denoted as $ \left \{ g_{i}^{p} \right \}_{i=0}^{\hat{M}} $ and $ \left \{ g_{i}^{c} \right \}_{i=0}^{\hat{M}} $, where $ \hat{M}\geq M $ denotes the number of groups. Consequently, the $j$-th neighborhood central anchor query $ q_{ij} $ in the $i$-th group is expressed as shown in \textcolor{red}{Equation}\eqref{pythagorean3}.
\begin{equation}
	q_{ij}  = Cat(p_{j} + g_{i}^{p},c_{j} + g_{i}^{c}) 
	\label{pythagorean3}.
\end{equation}

Regarding the representation of non-neighborhood central anchor queries, we assume that the distance between $p_{ij}'$ and $p_{ij}$ is $(\Delta x_{ij},\Delta y_{ij})$, which can be computed using \textcolor{red}{Equation}\eqref{pythagorean5}.

\begin{equation}
	\left\{\begin{matrix}
		& \Delta x_{ij} = \beta _{1}\ast a/2\\ 
		& \Delta y_{ij} = \beta _{2}\ast a/2\\
	\end{matrix}\right. 
	\label{pythagorean5},
\end{equation}
where $ \beta_{1} $ and $ \beta_{2} $ are random numbers in the interval $(-1,1) $.
Subsequently, according to \textcolor{red}{Equation}(\eqref{pythagorean1}--\eqref{pythagorean5}), the $j$-th non-neighborhood central anchor query $ q_{ij}' $ in the $i$-th group can be expressed as shown in \textcolor{red}{Equation}\eqref{pythagorean9}.
\begin{equation}
	q_{ij}'  = Cat((p_{j} + g_{i}^{p})+(\Delta x_{ij}, \Delta y_{ij}),c_{j} + g_{i}^{c}). 
	\label{pythagorean9}
\end{equation}

In summary, the neighborhood central anchor queries are denoted as $ Q = \left \{Q_{i} \right \}_{i=0}^{\hat{M}} = \left \{ \left \{ q_{ij} \right \}_{j=0}^{N} \right \}_{i=0}^{\hat{M}}$ and the non-neighborhood central anchor queries are denoted as $ Q' = \left \{Q_{i}' \right \}_{i=0}^{\hat{M}} = \left \{ \left \{ q_{ij}' \right \}_{j=0}^{N} \right \}_{i=0}^{\hat{M}}$ .

\subsection{GT Neighborhoods}
\begin{figure}[b]
	\centering
	\includegraphics[width=1\textwidth, trim=28cm 32cm 30cm 23cm, clip]{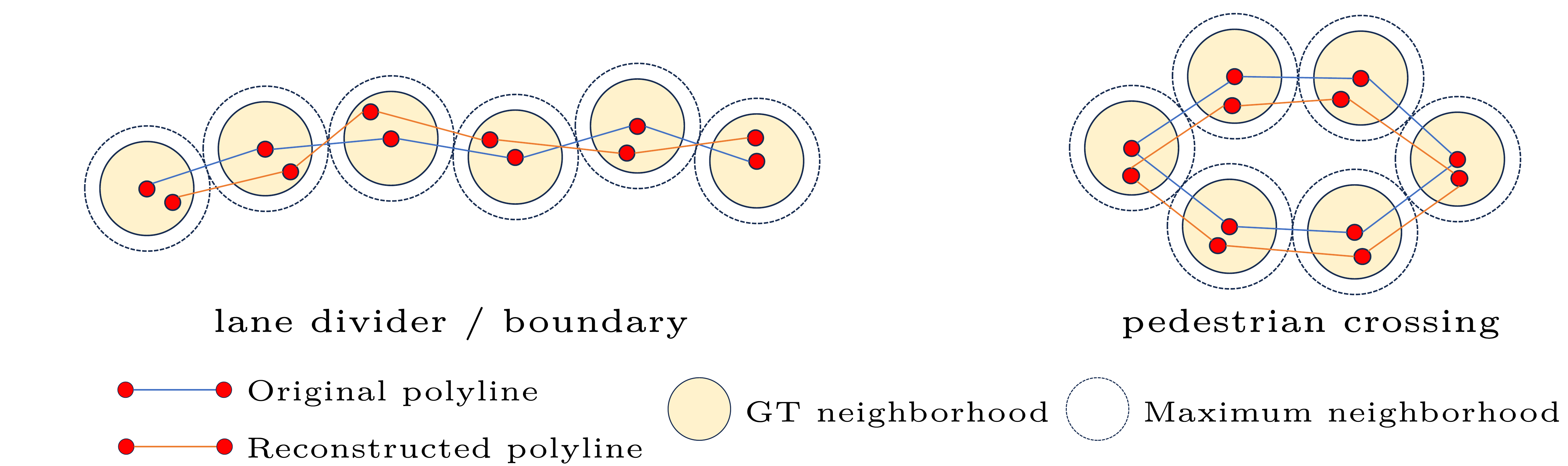}
	\caption{
		GT neighborhoods: We determine the maximum radius $r$ of the GT neighborhoods by half of the distance between vertices, and then further reduce the radius according to $\omega$, so as to maintain the overall shape feature of the map elements to the greatest extent. 
	}
	\label{fig4}
\end{figure}

However, ambiguities arise when both neighborhood central anchors and non-neighborhood central anchors are fitted to the same target points using identical content parts. To address this issue, we propose the GT neighborhoods, where non-central anchors are no longer fitted to target points but are instead fitted within the GT neighborhoods.
As illustrated in \textcolor{red}{Fig}\eqref{fig4}, we expect that the polyline ${V_{i}}'$ reconstructed from the points within the GT neighborhoods should undergo minimal change in shape compared to original polyline ${V_{i}}$. Therefore, determining the shape and size of the neighborhoods is crucial.

As the vertices of the GT polylines are uniformly sampled and consistent in number across different polylines, variations in distances between vertices may occur due to differences in polyline lengths. To address this, we define the neighborhood of each vertex as a circular region with a radius of $ r$, where $r$ is capped at half the distance between vertices.
Suppose the distance between vertices is $D$. Subsequently, the deviation range $(\Delta m_{i}, \Delta n_{i})$ of non-neighborhood central vertex $v_{j}'$ relative to neighborhood central vertex $v_{j}$ is determined by \textcolor{red}{Equation}\eqref{pythagorean7}.
\begin{equation}
	\left\{\begin{matrix}
		& r = \omega  \ast \left ( D/2 \right )\\ 
		& \Delta m_{i} = \beta _{1}\ast r\\ 
		& \Delta n_{i}= \beta _{2}\ast \sqrt{r^{2}- {\Delta m_{i}}^{2}}\\
	\end{matrix}\right.
	\label{pythagorean7},
\end{equation}
where $ \omega  \in (0,1] $, $ \beta_{1} $ and $ \beta_{2} $ are random numbers in the interval $(-1,1) $.

\subsection{GL-SA}

\begin{figure}[b]
	\centering
	\begin{minipage}[t]{0.48\textwidth}
		\centering
		\includegraphics[width=6cm,trim=11.8cm 8cm 9cm 5.5cm, clip]{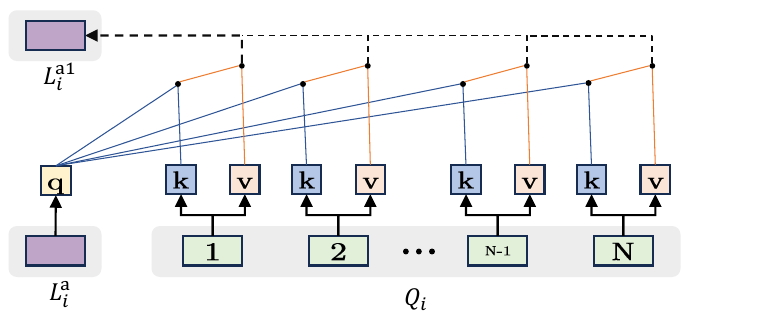}
		\caption{
			Local feature extraction: A single local query queries a group of anchor queries to extract local feature.
		}
		\label{fig5}
	\end{minipage}
	\hspace{0.11in}
	\begin{minipage}[t]{0.48\textwidth}
		\centering
		\includegraphics[width=6cm,trim=9cm 6.9cm 9cm 5.5cm, clip]{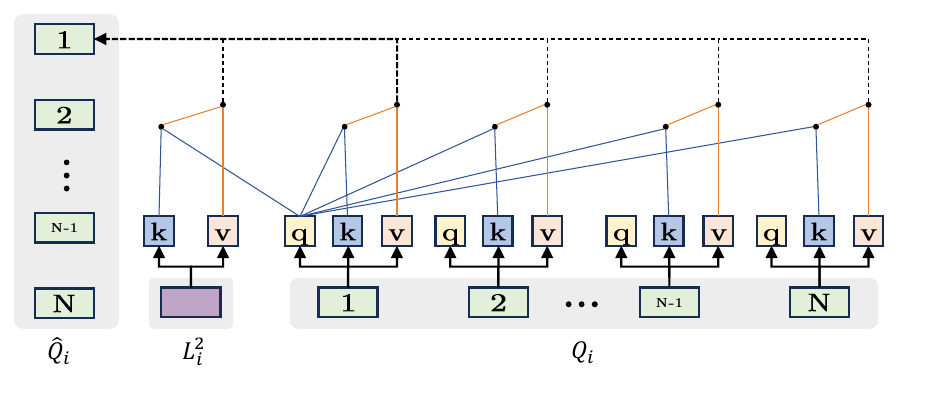}
		\caption{		
			Feature interaction within groups: In each group, each anchor query queries all queries to facilitate feature interaction within groups.
		}
		\label{fig6}
	\end{minipage}
\end{figure}
For simplicity, we analyze the implementation process and complexity of each step in the single-head GL-SA. The pseudocode of GL-SA can be found in \textcolor{red}{Alg}\eqref{algorithm}.
We define a set of learnable parameters $ L= \left \{ L_{i} \right \}_{i=0}^{\hat{M}} $ as local queries. The element $L_{i} $ is used to query the anchor queries $ Q_{i} $ to obtain the local feature of the $i$-th group.
Due to the introduction of anchor neighborhoods, the number of initial queries has doubled. To enhance the capability of aggregating feature using local queries, We partition $L_{i}$ into two parts, $L_{i}^{a}$ and $L_{i}^{b}$, doubling the dimensionality of the local queries compared to the anchor queries. Moreover, The feature aggregation method for both $L_{i}^{a}$ and $L_{i}^{b}$ is consistent.  
The experimental validation of the structural improvement of local queries is presented in \textcolor{red}{Section}\eqref{Ablation3}(3).

\SetKwComment{Comment}{\scriptsize/* }{ \scriptsize */}
\begin{algorithm}[t] \footnotesize
	\setstretch{1.2} 
	\caption{GL-SA}\label{algorithm}
	
	\KwData {Local queries $ L= \left \{ L_{i} \right \}_{i=0}^{\hat{M}} $, Anchor queries $Q = \left \{Q_{i} \right \}_{i=0}^{\hat{M}}=\left \{Cat(P_{i},C_{i}) \right \}_{i=0}^{\hat{M}}$} 
	
	\KwResult {Anchor queries $\hat{Q}$ for which the self-attention computation is completed}
	
	$q,k\leftarrow \left \{ C_{i} \right \}_{i=0}^{\hat{M}}+sinPE(\left \{ P_{i} \right \}_{i=0}^{\hat{M}})$ \Comment*[r]{ \scriptsize $ sinPE(\cdot ) $: Sinusoidal Position Encoding.}
	
	\ $v\leftarrow \left \{ C_{i} \right \}_{i=0}^{\hat{M}}$ \Comment*[r]{\scriptsize $v\in  \mathbb{R}^{b*(\hat{M}*N)*d}$, $b$: batch size}
	
	$L^{a}, L^{b}\leftarrow split(L,dim=2)$ \Comment*[r]{\scriptsize $L^{a}, L^{b}\in  \mathbb{R}^{\hat{M}*b*d}$}
	
	{\bf{(1) Local feature extraction}}
	
	$k^{1},v^{1} \leftarrow RLinear(k,v)$ \Comment*[r]{ \scriptsize RLinear(.): reconstructing the shape of the output from the linear layer. $ k^{1},v^{1}\in  \mathbb{R}^{(\hat{M}*b)*N*d}$}
	
	$q^{L^{a}},q^{L^{b}} \leftarrow RLinear(L^{a}, L^{b})$ \Comment*[r]{ \scriptsize $q^{L^{a}},q^{L^{b}}\in  \mathbb{R}^{(\hat{M}*b)*1*d}$}			
	
	$L^{a1} \leftarrow LF(q^{L^{a}},k^{1}, v^{1} )$\Comment*[r]{\scriptsize LF(.): local feature extraction. }
	
	$L^{b1} \leftarrow LF(q^{L^{b}}, k^{1}, v^{1} )$\Comment*[r]{\scriptsize$L^{a1}, L^{b1}\in  \mathbb{R}^{\hat{M}*b*d}$}
	
	{\bf{(2) Feature interaction among groups}}
	
	$ Q_{r}  \leftarrow reshape( \left \{ C_{i} \right \}_{i=0}^{\hat{M}}+sinPE(\left \{ P_{i} \right \}_{i=0}^{\hat{M}}))$ \Comment*[r]{\scriptsize$Q_{r}\in  \mathbb{R}^{\hat{M}*N*b*d}$}
	
	$ Q_{m} \leftarrow mean(Q_{r},dim=1)$ \Comment*[r]{\scriptsize$Q_{m}\in  \mathbb{R}^{\hat{M}*b*d}$}
	
	$ L^{2'} \leftarrow mean(L^{a1} + L^{b1}) + Q_{m}$; \Comment*[r]{\scriptsize$L^{2'}\in  \mathbb{R}^{\hat{M}*b*d}$}
	
	$ L^{2} \leftarrow van.(L^{2'})$ \Comment*[r]{ \scriptsize $ van.(\cdot ) $: vanilla self-attention.}
	
	{\bf{(3) Feature interaction within groups}}
	
	$ q^{2}, k^{2},v^{2} \leftarrow RLinear(q, k, v)$ \Comment*[r]{\scriptsize $q^{2}, k^{2},v^{2}\in  \mathbb{R}^{(\hat{M}*b)*N*d}$}
	
	$ k^{L}, v^{L} \leftarrow RLinear(L^{2} )$ \Comment*[r]{\scriptsize $ k^{L}, v^{L}\in  \mathbb{R}^{(\hat{M}*b)*1*d}$}	
	
	$ k^{LQ}, v^{LQ} \leftarrow Cat((k^{L}, k^{2}),(v^{L}, v^{2}))$ \Comment*[r]{\scriptsize $ k^{LQ},v^{LQ}\in  \mathbb{R}^{(\hat{M}*b)*(N+1)*d}$}
	
	$\hat{Q} \leftarrow  FI(q^{2}, k^{LQ}, v^{LQ} )$ \Comment*[r]{\scriptsize FI(.): feature interaction within groups. }
	
	\Return{$\hat{Q}$}
	
\end{algorithm}

\subsubsection{(1) Local Feature Extraction.} 
To simplify further, we analyze the process of local feature extraction by $L_{i}^{a}$ in a group of anchor queries.
In \textcolor{red}{Fig}\eqref{fig5}, $Q_{i}$ generates $k_{i}^{1}$ and $v_{i}^{1}$ by a linear layer, while $L_{i}^{a}$ generates $q_{L}$ by another linear layer. So, we have $ q_{L} \in  \mathbb{R}^{1*d} $, $ k_{i}^{1} \in \mathbb{R}^{N*d} $ and  $v_{i}^{1} \in \mathbb{R}^{N*d} $. 
Subsequently, $q_{L}$ queries $k_{i}^{1}$ to obtain the attention matrix $A_{i}^{1}$, with the computational complexity of $ A_{i}^{1} = q_{L}\ast (k_{i}^{1})^{T}$ is $O (N * d) $.
Afterwards, the computational complexity of $softmax(A_{i}^{1}) $ is $O(N)$. 
Upon applying softmax, $ A_{i}^{1}$ is multiplied by $v_{i}^{1}$ to obtain $L_{i}^{a1}$, resulting in a computational complexity of  $O(N *d)$. So, the computational complexity of {\bf{(1)}} is $ O_{1} = O(N*d*\hat{M}) $.

\subsubsection{(2) Feature Interaction among Groups.}
Firstly, compute the mean $Q_{m}$ of $(\left \{ C_{i} \right \}_{i=0}^{\hat{M}} + sinPE(\left \{ P_{i} \right \}_{i=0}^{\hat{M}})) $. Next, calculate the mean of the sum of $L^{a1}$ and $L^{b1}$, and add $Q_{m}$ to obtain the local feature $L^{2'}$. $Q_{m}$ supplements the local feature $L^{2'}$, similar to residual connections. Then $L^{2'}$ is fed into vanilla self-attention to get $L^{2} $,  thereby realizing the feature interaction among groups. Given $L^{2'}\in R^{\hat{M}*d} $, the computational complexity of  {\bf{(2)}} is $ O_{2} = O(\hat{M}^{2}*d) $.

\subsubsection{(3) Feature Interaction within Groups.}
The elements of $L^{2}$ are assigned to the initial groups, and in each group, anchor queries query all the queries including the local queries to realize the feature interaction within groups.
In \textcolor{red}{Fig}\eqref{fig6}, the keys and values generated by $L_{i}^{2}$ and $Q_{{i}}$  respectively combine to form $k_{i}^{LQ}$ and $v_{i}^{LQ}$, so $k_{i}^{LQ} \in \mathbb{R}^{(N+1)*d} $ and $ v_{i}^{LQ}\in  \mathbb{R}^{(N+1)*d} $.
Next, $q_{i}^{2}\in  \mathbb{R}^{N*d}$ generated by $Q_{i}$ queries $k_{i}^{LQ}$ to obtain the attention matrix $A_{i}^{2}$,
with the computational complexity of $ A_{i}^{2} = q_{i}^{2}\ast k_{i}^{LQ} $ is $ O(N*(N+1)*d) $.
Afterwards, the computational complexity of $softmax(A_{i}^{2}) $ is $O(N*(N+1))$. 
Following the softmax application, $A_{i}^{2}$ is multiplied by $v_{i}^{LQ}$ to obtain $\hat{Q}_{i}$, resulting in a computational complexity of $O(N*(N+1)*d)$. Consequently, the computational complexity of  {\bf{(3)}} is $ O_{3} =  O(N*(N+1)*d*\hat{M}) $.

In summary, the computational complexity of GL-SA is $ O_{GL} = O_{1} + O_{2} + O_{3} $. Meanwhile, for vanilla self-attention, the computational complexity is $ O_{van.} = O(((\hat{M}*N)^{2})*d) $. 
To compare GL-SA and vanilla self-attention, we use the scaling factor $\partial$, calculated as shown in \textcolor{red}{Equation}\eqref{pythagorean10}.
Therefore, the computational complexity of GL-SA is markedly reduced. In the experimental section, we validate the efficiency of GL-SA by comparing the memory consumption with vanilla self-attention during the training process.
\begin{equation}
	\partial  = \frac{O_{GL}}{O_{van.}} \approx  \frac{1}{\hat{M}} + \frac{1}{N^{2}} \ll 1 .
	\label{pythagorean10} 
\end{equation}

\subsection{Matching and Supervise}

The structure of EAN-MapNet is divided into a primary part and an auxiliary part, both of which are separately matched and supervised, with the matching and supervision methods inspired by MapTR\cite{liao2022maptr}.
Assuming the loss of the primary part is denoted as $L_{center}$ and the loss of the auxiliary part is denoted as $L_{non-center}$. 
Referring to MapTRv2\cite{liao2023maptrv2}, We enhance the supervision of EAN-MapNet by incorporating segmentation-based loss $L_{seg.}$ and depth estimation loss $L_{depth}$. Consequently, the loss function of EAN-MapNet is shown \textcolor{red}{Equation}\eqref{pythagorean8}. 
For BEV feature, $L_{seg.}$ loss enhances the feature representation of map elements regions, while $L_{depth}$ bolsters the representation of depth information.

\begin{equation}
	L = \lambda_{1} L_{center} + \lambda_{2} L_{non-center} + \lambda_{3} L_{seg.} + \lambda_{4} L_{depth}. \label{pythagorean8} 
\end{equation}

\section{Experiments}
\subsection{Dataset and Metric}
We evaluate the performance of the EAN-MapNet using the nuScenes dataset\cite{caesar2020nuscenes}. The model is trained with images captured by 6 surround-view cameras. The predicted HD map range within a single frame spans $[-15m, 15m]$ along the X-axis and $[-30m, 30m]$ along the Y-axis in the ego-vehicle coordinate system. The predicted map elements encompass categories such as pedestrian crossings, lane dividers, and boundaries.

We matche the predicted and GT using three thresholds of Chamfer distance $\tau = {0.5, 1.0, 1.5}$, and then calculated the map construction quality $AP_{\tau}$ for each threshold. Subsequently, the average $AP_{m}$ of $ \left \{ AP_{\tau } \right \}_{\tau }^{\left \{ 0.5,1.0,1.5 \right \}}$ is  utilized as a criterion for evaluating the model's performance.

\subsection{Implementation Details}
EAN-MapNet utilizes GKT\cite{chen2022efficient} as the BEV feature generation network. Furthermore, to ensure a fair comparison with MapTRv2, EAN-MapNet switches to employing LSS\cite{philion2020lift} as the BEV feature generation network after incorporating depth estimation loss and segmentation-based loss. EAN-MapNet take images as input, with the image size set to (480, 800).
The optimizer employs AdamW with an initial learning rate set to  $2.5 \times  10^{-4}$. The default training schedule is 24 epochs , with a batch size of 4. 
In \textcolor{red}{Section}\eqref{Ablation1}(1) and \textcolor{red}{Section}\eqref{Ablation2}(2), the grouping number of initial query units is set to 50, and 1 NVIDIA A40 is utilized with a weight decay of 0.01. For all other conditions, the grouping number of initial query units is set to 100, and experiments are conducted using 1 NVIDIA GeForce RTX 3090 with the weight decay of $1.25 \times  10^{-3}$.
In the loss function, $\lambda_{0} = \lambda_{1} = \lambda_{2} = \lambda_{3} = 1$. Regarding the neighborhood configuration, when local queries are not improved, $\omega  = 0.25$ and $a = 0.55m$; otherwise, $\omega  = 0.2$ and $a = 0.5m$.

\subsection{Comparisons with State-of-the-art Methods}

\begin{table*}[tb] \footnotesize
	\setlength{\tabcolsep}{6pt}
	\renewcommand{\arraystretch}{1.5}
	\centering
	\caption{
		Comparisons with SOTAs. "-" indicates that the result is not available. The $^{\bigstar}$ indicates the inclusion of both depth estimation loss and segmentation loss as supervision methods.
		For fairness, all methods employ ResNet\cite{he2016deep} as the backbone and solely take images as input. Simultaneously, on an RTX 3090, we maintain an image size of (480, 800) for FPS testing and retain a batch size of 4 for memory consumption testing.
	}
	\label{table}
	\begin{tabular}{c|c|cccc|c|c}
		\hline\noalign{\smallskip}
		\multirow{2}*{Method}  &  \multirow{2}*{Epoch}&   \multicolumn{4}{c|}{AP$\uparrow$ }  & \multirow{2}*{FPS $\uparrow$} & \multirow{2}*{GPU mem. $\downarrow$}   \\
		&        &       ped. & div. & bou. & mean  &   &     \\
		\noalign{\smallskip}
		\hline
		\noalign{\smallskip}
		\multirow{1}*{VectorMapNet\cite{liu2023vectormapnet}}   &     110&  36.1&  47.3&  39.3&  40.9& 1.5 &20237M     \\
		
		\multirow{1}*{BeMapNet\cite{qiao2023end}}    &    30&  57.7&  62.3&  59.4&  59.8&  -&-    \\
		
		\multirow{1}*{PivotNet\cite{ding2023pivotnet}}    &    24& 56.2&  56.5&  60.1&  57.6&   -&-     \\
		
		\noalign{\smallskip}
		\hline
		\noalign{\smallskip}
		\multirow{1}*{MapTR\cite{liao2022maptr}}    &   24&  46.3&  51.5&  53.1&  50.3& {\bf{17.2}}&{\bf{10152M}}     \\
		
		\multirow{1}*{EAN-MapNet}   &    24&  {\bf{55.9}}&  {\bf{61.7}}&  {\bf{61.5}}&  {\bf{59.7}}&  14.9&10838M    \\

		\noalign{\smallskip}
		\hline
		\noalign{\smallskip}
		
		\multirow{1}*{MapTRv2\cite{liao2023maptrv2}}    &    24&  59.8&  {\bf{62.4}}&  62.4&  61.5&    {\bf{13.3}}&19299M   \\
		
		\multirow{1}*{EAN-MapNet$^{\bigstar }$}    &    24&  {\bf{60.5}}&  62.1&  {\bf{66.3}}&  63.0&   12.6&\bf{{11101M}}   \\
		\hline
	\end{tabular}
\end{table*}	

\begin{figure}[tb]
	\centering
	\includegraphics[width=1\textwidth, trim=0cm 5cm 0cm 0cm, clip]{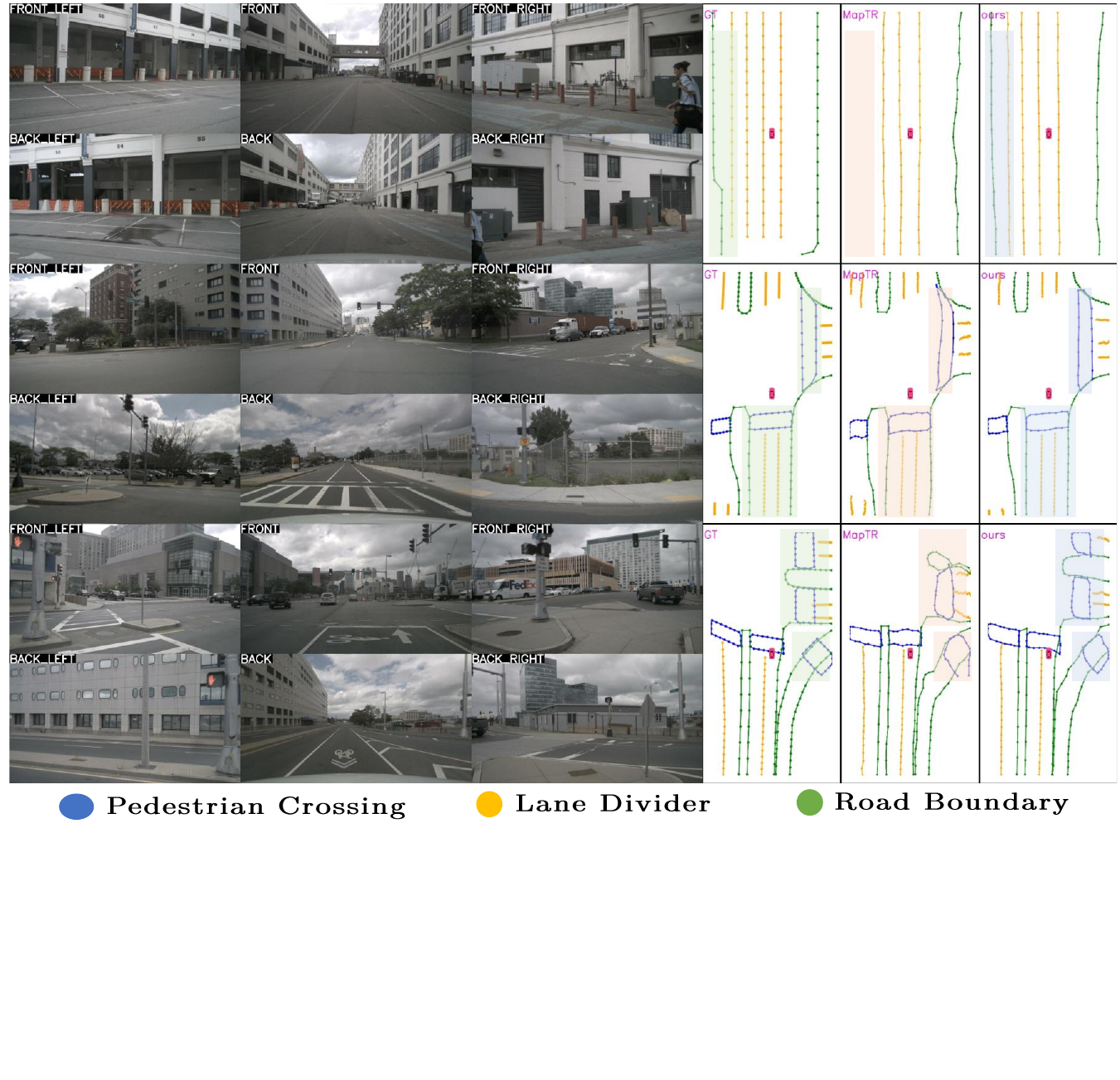}
	\caption{
		Comparison with baseline on qualitative visualization. The visualizations demonstrate the superiority and robustness of EAN-MapNet in the task of detecting map elements.
	}
	\label{fig7}
\end{figure}

As shown in\textcolor{red}{Table}\eqref{table}, we compare EAN-MapNet with state-of-the-art methods. 
Compared to MapTR\cite{liao2022maptr}, EAN-MapNet demonstrates a significant improvement of 9.4 mAP. Additionally, despite doubling the number of initial queries in EAN-MapNet compared to MapTR, the inference speed decreases by only 2.2 fps. Furthermore, with identical segmentation-based loss and depth estimation loss, EAN-MapNet achieves a 1.5 mAP improvement over MapTRv2\cite{liao2023maptrv2}, while also reducing memory consumption by 8198M during training. Additionally, EAN-MapNet outperforms PivotNet\cite{ding2023pivotnet} by 2.6 mAP. In comparison to BeMapNet\cite{qiao2023end}, EAN-MapNet significantly outperforms it with a margin of 3.2 mAP.
Simultaneously, we performed qualitative visualization in \textcolor{red}{Fig}\eqref{fig7}.

\subsection{Ablation Study}
\label{Ablation}

\begin{table}[tb]
	\centering
	\begin{minipage}[t]{0.48\textwidth} \footnotesize
		\centering
		\setlength{\tabcolsep}{8pt}
		\renewcommand{\arraystretch}{1.75}
		\caption{
			Effectiveness of GL-SA with anchor queries.}
		\label{table1}
		\begin{tabular}{c|cc}
			\hline\noalign{\smallskip}
			&   mAP  &GPU mem.      \\
			\noalign{\smallskip}
			\hline
			\noalign{\smallskip}
			vanilla queries & 52.4    & 10152M    \\
			anchor queries  & 53.6    & 10235M    \\
			GL-SA           & 54.9    & 8022M     \\
			\hline
		\end{tabular}
	\end{minipage}
	\hspace{0.11in}
	\begin{minipage}[t]{0.48\textwidth} \footnotesize
		\centering
		\setlength{\tabcolsep}{10pt}
		\renewcommand{\arraystretch}{1.2}
		\caption{
			Effectiveness of each module in GL-SA, $\mathbb{L}$ denotes the local queries, and $\mathbb{M}$ represents the mean of anchor queries.}
		\label{table2}
		\begin{tabular}{cc|cc}
			\hline\noalign{\smallskip}
			$\mathbb{L}$ & $\mathbb{M}$ & \scriptsize mAP & \scriptsize GPU mem.  \\
			\noalign{\smallskip}
			\hline
			\noalign{\smallskip}
			$\times$   & $\times$   & 53.0  &  7922M\\ 			
			\checkmark & $\times$   & 53.9  &  8022M\\ 
			\checkmark & \checkmark & 54.9  &  8022M\\
			\hline
		\end{tabular}
		
	\end{minipage}
\end{table}

\subsubsection{(1) Effectiveness of GL-SA with Anchor Queries.}
\label{Ablation1}
As shown in \textcolor{red}{Table}\eqref{table1}, our defined anchor queries lead to a 1.2 mAP improvement compared to using vanilla queries. Subsequently, After incorporating the use of GL-SA on top of setting anchor queries, there is a reduction in memory consumption by 2213M, accompanied by a corresponding 1.3 mAP improvement. This validates that GL-SA can significantly reduce computational complexity while fully achieving feature interaction among queries.

\subsubsection{(2) Effectiveness of Each Module in GL-SA.}
\label{Ablation2}
As shown in \textcolor{red}{Table}\eqref{table2}, the introduction of local queries leads to a mere 100 M increase in memory consumption, while yielding a 0.9 mAP improvement. This suggests that local queries can effectively serve as media for feature interaction among queries from different groups. Moreover, the impact of $\mathbb{M}$ is apparent, leading to a 1.0 mAP improvement without incurring additional memory overhead. This indicates that the supplementation of local feature by $\mathbb{M}$ is significant.

\begin{table}[tb]
	\centering
	\setlength{\tabcolsep}{24pt}
	\renewcommand{\arraystretch}{1.2}
	\caption{
		Effectiveness of the anchor neighborhoods.}
	\label{table3}
	\begin{tabular}{c|c|c}
		\hline\noalign{\smallskip}
		Index&                              Method&   mAP \\
		\noalign{\smallskip}
		\hline
		\noalign{\smallskip}
		(a)&                            baseline&   56.2\\
		(b)&        (a) $+$ anchor neighborhoods&   56.7\\           					    
		(c)&            (b) $+$ GT neighborhoods&   58.7\\           					    
		(d)&     (c) $+$ improved local queries &   59.7\\           
		\hline
	\end{tabular}
\end{table}

\subsubsection{(3) Effectiveness of Anchor Neighborhoods.}
\label{Ablation3}

We conducte four sets of experiments: (a) The baseline employs GL-SA and anchor queries, (b) introducing anchor neighborhoods on top of (a), (c) introducing GT neighborhoods on top of (b), and (d) utilizing improved local queries on top of (c). The results are shown in \textcolor{red}{Table}\eqref{table3}.
Comparing (a) and (b), the use of anchor neighborhoods results in only a 0.5 mAP increase. This occurs due to ambiguity when both neighborhood central anchors and non-neighborhood central anchors are fitted to the same target points using identical content parts. Therefore, with the introduction of GT neighborhoods in (c), the network demonstrates a substantial improvement of 2.5 mAP. Furthermore, when comparing (d) with (c), there is an increase of 1.0 mAP, indicating that the improved structure of the local queries can significantly enhance the ability to aggregate local feature.

\subsubsection{(4) Additional Evidence Confirming the Effectiveness of Anchor Neighborhoods.}
\label{Ablation4}
As shown in \textcolor{red}{Table}\eqref{table4}, when the positions of the non-neighborhood central anchors are not confined within anchor neighborhoods but are randomly distributed on the BEV plane, the experimental result is 1.4 mAP higher than that of "anchor neighborhoods" without the introduction of GT neighborhoods. However, with the introduction of GT neighborhoods to alleviate fitting ambiguities, "anchor neighborhoods" surpasses "random anchors" by 0.8 mAP. This indicates that when non-neighborhood central anchors are confined within anchor neighborhoods, they exhibit stronger positional correlation with neighborhood central anchors, thereby better assisting in fitting the neighborhood central anchors to the target points after alleviating fitting ambiguities.

\begin{table}[tb]
	\centering
	\setlength{\tabcolsep}{16pt}
	\renewcommand{\arraystretch}{1.2}
	\caption{
		Effectiveness of the anchor neighborhoods. When non-neighborhood central anchors are within the neighborhoods, labeled as "anchor neighborhoods", and without restrictions on the neighborhoods, labeled as "random anchors".}
	\label{table4}
	\begin{tabular}{c|c|c}
		\hline\noalign{\smallskip}
		& GT neighborhoods & \scriptsize mAP   \\
		\noalign{\smallskip}
		\hline
		\noalign{\smallskip}
		\multirow{2}*{anchor neighborhoods}& $\times$   & 56.7 \\
		~& \checkmark & 58.7      \\  
		\noalign{\smallskip}
		\hline
		\noalign{\smallskip}              
		\multirow{2}*{random anchors} & $\times$   & 58.1  \\ 
		~& \checkmark & 57.9 \\
		\hline
	\end{tabular}
\end{table}

\section{Conclusions}
In this paper, we propose EAN-MapNet, which incorporates the physical positional information of anchor neighborhoods into its initial query units. Furthermore, we introduce GL-SA, a method that efficiently facilitates ample feature interaction among anchor queries. On the nuScenes dataset, we demonstrate that non-neighborhood central anchors effectively assist in accurately fitting neighborhood central anchors to target points. As for GL-SA, experimental results demonstrate its ability to facilitate more effective feature interaction among queries while greatly reducing computational complexity. Additionally, compared with other works, EAN-MapNet achieves a state-of-the-art performance. 
Although we have obtained promising experimental results, we only utilize a single randomly distributed non-neighborhood central anchor in each query unit, which limits the utilization of positional information from anchor neighborhoods. Thus, in future research, we can explore incorporating multiple non-neighborhood central anchors to further enhance the structure of anchor query units. Meanwhile, we hope that GL-SA can provide valuable insights into the task of labels presentation grouping distribution status.

\clearpage
%
%
\bibliographystyle{unsrt}
\bibliography{EAN-MapNet}
\end{document}